\documentclass[letterpaper]{article} 
\usepackage[]{aaai23}  
\usepackage{times}  
\usepackage{helvet}  
\usepackage{courier}  
\usepackage[hyphens]{url}  
\usepackage{graphicx} 
\usepackage{natbib}  
\usepackage{caption} 
\usepackage{subfigure}
\usepackage{algorithm}
\usepackage{algorithmic}
\usepackage{amsmath}
\usepackage{amssymb}
\usepackage{mathtools}
\usepackage{amsthm}
\usepackage{booktabs}
\usepackage{multirow}
\usepackage[normalem]{ulem}
\useunder{\uline}{\ul}{}
\usepackage{xcolor}
\definecolor{degreen}{RGB}{0,170,85}
\usepackage{newfloat}
\usepackage{listings}
\DeclareCaptionStyle{ruled}{labelfont=normalfont,labelsep=colon,strut=off} 
\floatstyle{ruled}
\newfloat{listing}{tb}{lst}{}
\floatname{listing}{Listing}
\title{Protein 3D Graph Structure Learning for Robust Structure-based\\ Protein Property Prediction}
\author{
    Yufei Huang\equalcontrib \textsuperscript{\rm 1,2}, Siyuan Li\equalcontrib \textsuperscript{\rm 1,2}, Lirong Wu\textsuperscript{\rm 1,2}, Jin Su\textsuperscript{\rm 1,2}, Odin Zhang\textsuperscript{\rm 2}, \\ Haitao Lin\textsuperscript{\rm 1,2}, Jingqi Qi\textsuperscript{\rm 3}, Zihan Liu\textsuperscript{\rm 1,2}, Zhangyang Gao\textsuperscript{\rm 1,2}, Yuyang Liu\textsuperscript{\rm 2}, Jiangbin Zheng\textsuperscript{\rm 1,2}, \\Stan Z. Li\textsuperscript{\rm 2}\thanks{Corresponding Author}
}
\affiliations{

    \textsuperscript{\rm 1} Zhejiang University, Hangzhou\\
    \textsuperscript{\rm 2} AI Lab, Research Center for Industries of the Future, Westlake University\\
    {huangyufei, lisiyuan, sujin, wulirong, linhaitao, zhengjiangbin, Stan.ZQ.Li}@westlake.edu.cn, \\liuyuyang@gmail.com, jingqq@uw.edu
}

\usepackage{bibentry}

\begin{document}

\maketitle

\begin{abstract}
Protein structure-based property prediction has emerged as a promising approach for various biological tasks, such as protein function prediction and sub-cellular location estimation. The existing methods highly rely on experimental protein structure data and fail in scenarios where these data are unavailable. Predicted protein structures from AI tools (e.g., AlphaFold2) were utilized as alternatives. However, we observed that current practices, which simply employ accurately predicted structures during inference, suffer from notable degradation in prediction accuracy. While similar phenomena have been extensively studied in general fields (e.g., Computer Vision) as model robustness, their impact on protein property prediction remains unexplored. In this paper, we first investigate the reason behind the performance decrease when utilizing predicted structures, attributing it to the structure embedding bias from the perspective of structure representation learning. To study this problem, we identify a Protein 3D Graph Structure Learning Problem for Robust Protein Property Prediction (PGSL-RP3), collect benchmark datasets, and present a protein \underline{\textbf{S}}tructure embedding \underline{\textbf{A}}lignment \underline{\textbf{O}}ptimization framework (\textbf{\texttt{SAO}}) to mitigate the problem of structure embedding bias between the predicted and experimental protein structures. Extensive experiments have shown that our framework is model-agnostic and effective in improving the property prediction of both predicted structures and experimental structures. The benchmark datasets and codes will be released to benefit the community.
\end{abstract}

\section{Introduction}
Proteins are workhorses of the cell, involved in various biological processes, such as immune response and DNA replication. Understanding the properties of proteins is important for deciphering the mystery of life~\cite{PDBminer,hu2022protein} and treating various diseases~\cite{rare_diseases, zheng2023lightweight}. As most protein properties are governed by their folded structures, protein structure-based property prediction has emerged as a promising approach for various biological tasks, such as protein function prediction~\cite{huang2023data}, sub-cellular location estimation~\cite{zhang2022protein}, structure-based drug design~\cite{lin2022diffbp,lin2023functionalgroupbased}, and antibody design~\cite{kong2022conditional}. 

\begin{figure}[!tbp]
    \begin{center}
    \hspace{-1em}
        \subfigure[Performance Drop]{\includegraphics[width=0.55\linewidth]{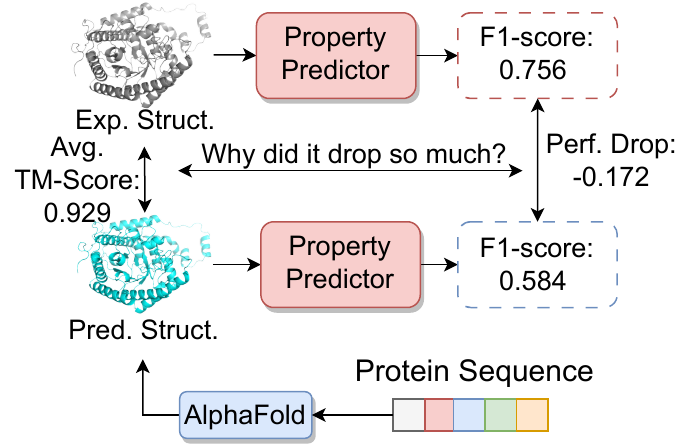}\label{fig:1a}}
        \hspace{-1em}
        \subfigure[Embedding Bias]{\includegraphics[width=0.5\linewidth]{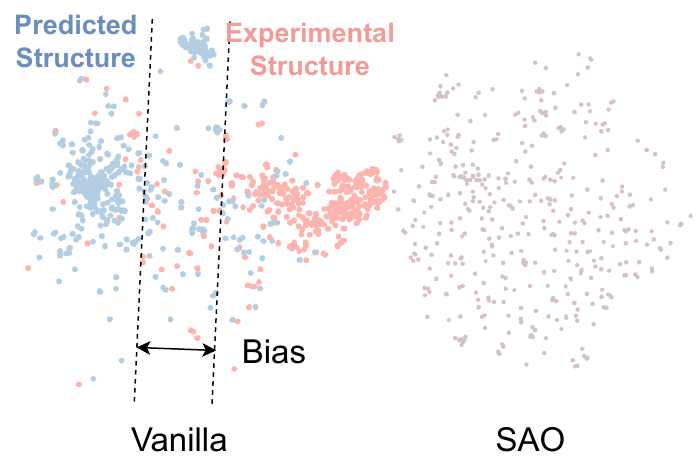}\label{fig:1b}}
    \hspace{-1.5em}
    \end{center}
    \vspace{-0.5em}
    \caption{The illustration of our finding problem. (a) Performance drop on EC task. TM-score is a widely used metric for structure prediction accuracy
            (b) The illustration of vanilla and our learned protein structure embedding of a subset of GOCC dataset by t-SNE. The Red is experimental structure embedding, and the blue is predicted structure embedding. There is a clear bias between the predicted and experimental structures in embeddings from vanilla encoder but embeddings from \texttt{SAO}-pretrained encoder tend to be smoother and bias-free.}
    \label{fig:1}
    \vspace{-2em}
\end{figure}

The existing methods' reliance on experimental protein structures poses a challenge when such structures are unavailable. Predicted protein structures from tools like AlphaFold2 have been utilized as alternatives. However, for various structure-based protein property prediction methods, even if these predicted structures are \textbf{accurate}, using them during inference also results in a notable decrease in property prediction accuracy. (as illustrated in Fig.\ref{fig:1a}). This suggests that there is a deeper factor behind the accuracy of predicted structures that misleads structure-based predictors. Similar phenomena also occur in other fields where new samples with small and hidden differences from the original ones can easily fool networks in making predictions for downstream tasks. In Computer Vision, when we apply small and invisible perturbations to a panda picture, the neural network misclassifies it to gibbon~\cite{goodfellow2015explaining}. Graph Neural Network is also known to be vulnerable to small perturbations like adding or deleting a few edges~\cite{GSL}. While developing robust algorithms(e.g., Graph Structure Learning, GSL~\cite{UGSL, GSL}) to resist permutations has been well studied in general domains, model robustness in structure-based protein property predictions remains unexplored.

Therefore, in this paper, we investigate the decrease in prediction accuracy when using accurately predicted protein structures. It is attributed to the \emph{\textbf{structure embedding bias}} of predictors from the perspective of structure representation learning, i.e., a distribution gap between embedding of accurately predicted structure and that of experimental structure as shown in Fig.\ref{fig:1b}. We further formulate this problem as Protein 3D Graph Structure Learning for Robust Protein Property Prediction (\emph{\textbf{PGSL-RP3}}). Improved predicted structures don't necessarily lead to better prediction results due to the structure embedding bias; the strategy of structure refinement, i.e., further improving the structure prediction accuracy, can't completely alleviate the performance decrease.

To address these issues, we propose to align the representation of the predicted structure to that of the experimental structure rather than directly improving the similarity (e.g., structure prediction accuracy) between predicted and experimental structures in data space. To achieve this, we present the protein \underline{\textbf{S}}tructure embedding \underline{\textbf{A}}lignment \underline{\textbf{O}}ptimization framework (\texttt{\textbf{SAO}}). In \texttt{SAO}, we create pairs of predicted and experimental structures and train the encoder to align the predicted structure representation with the corresponding experimental structure representation using a bootstrap and denoising approach. One advantage of our alignment-based framework is its ability to leverage paired information from both predicted and experimental structures, resulting in improved representation learning performance compared to using a single data source alone. Additionally, our framework can utilize low-precision predicted structures that are often overlooked, further enhancing its effectiveness. 

Our contributions can list as follows:
\begin{itemize}
\item We identify and formulate the problem of predicted structure embedding bias as Protein 3D Graph Structure Learning. To help further solve the problem, we collected relevant predicted structures and designed a comprehensive benchmark test. Datasets and codes will be released to benefit the community.
\item We propose the protein  \underline{\textbf{S}}tructure embedding \underline{\textbf{A}}lignment \underline{\textbf{O}}ptimization framework (SAO) to alleviate the structure embedding bias.
\item We conduct extensive experiments in our designed benchmark test. The results show our superior performance over various baselines and the ability to improve the property prediction of both predicted structures and experimental structures.
\end{itemize}

\section{Related Work}
\noindent \textbf{Protein Structure Representation Model.}
Graph neural networks~\cite{NEURIPS2022_KRD, wu2023extracting} were applied to protein structure representation~\cite{baldassarre2021graphqa,gligorijevic2021structure}, with promising results in protein structure design~\cite{ingraham2019generative,dauparas2022robust}. With the rise of geometric deep learning~\cite{lin2023nonequispaced, wu2022automated, wu2023quantifying}, Equivariant Neural Networks(ENN) began to be applied to protein representation, allowing us to directly learn protein structures end-to-end and achieve more powerful results~\cite{hsu2022learning,jing2020learning,wu2022atomic}. Inspired by AlphaFold~\cite{jumper2021highly}, many works have tried to combine sequence and structure representation, expecting that co-modeling can combine the advantages of both modalities~\cite{lin2022language,mansoor2021toward,you2022cross, wu2023extracting}.

\noindent \textbf{Protein Data Mining with Predicted Structure.}
With the introduction of AlphaFold2, many unknown structures have been solved predictively, which has fuelled the enthusiasm to transform large amounts of predicted structure into valuable knowledge~\cite{miningprotein, PDBminer}. Some progress has been made in certain areas~\cite{AF2kinase,rare_diseases,PDBminer}. In particular, areas with less mainstream attention, such as rare disease research, are expected to benefit greatly from AlphaFold2's predicted structures~\cite{rare_diseases}. At the same time, there is also a growing interest in the quality of knowledge generated by predictive structures, and there are now a number of relevant evaluation studies~\cite{miningprotein, PDBminer, af_mutatiom_eval}. However, current studies still focus on the accuracy of the prediction and ignore the structure embedding bias that the predicted structure itself may carry. EquiPPIS~\cite{EquiPPIS} has considered the problem of generalization to predicted structures and designed an equivariant neural network-based model for a specific task to alleviate this issue. We go further to reveal the nature of the problem and formulate it as the PGSL problem, thus proposing a more efficient and universal framework for its solution. More related work can be referred to the Appendix A.
\section{Preliminaries}
\noindent \textbf{Notions} Protein data can be modeled at multiple levels: sequence, amino acid level, full atom level, etc. Here we model proteins uniformly as an Attributed Relational Graph: $\mathcal{G} = (\mathcal{V}, \mathcal{E}, \mathcal{N}, \mathcal{R})$, where $\mathcal{V}$ represents the ordered set of graph nodes (can be amino acids or atoms) and $\mathcal{E} \in \mathcal{V} \times \mathcal{V}$ represents the corresponding set of edges connecting the nodes (some relationship between nodes, e.g., distance less than 4 \AA). Every vertex $v\in\mathcal{V}$ in $\mathcal{G}$ can have both scalar and vector attributes $\mathbf{n}_v=(S_v, V_v) \in \mathcal{N}$, where $S_v \in \mathbb{R}^S$ and $V_v \in \mathbb{R}^{3\times V}$. Similarly, each edge $e \in \mathcal{E}$ have attributes $\mathbf{r}_v=(S_e, V_e) \in \mathcal{R}$, where $S_e \in \mathbb{R}^N$ and $V_e \in \mathbb{R}^{3\times T}$. $\mathcal{G}$ can contain empty sets. When the sets $\mathcal{E}$ and $\mathcal{R}$ are empty sets, $\mathcal{G}$ degenerates to a single sequence representation. Furthermore, if $\mathcal{N}$ contains only amino acid composition, $\mathcal{G}$ degenerates to the amino acid sequence.

\noindent \textbf{PGSL for Protein Property Prediction.}
The aim of structure-based protein property prediction is to predict (classify or regress) the property of a protein given its structure. The Protein 3D Graph Structure Learning problem (PGSL) extends protein structures from experimental to predicted structures. Its goal is to enable protein structure representation models to align predicted structures to experimental structures, thus enabling representation models to better handle large numbers of predicted protein structures and improve humanity’s understanding of unknown proteins. Specifically for Protein Property Prediction, PGSL for Robust Protein Property Prediction (PGSL-RP3) requires that structure-based protein property prediction models can make correct annotations on predicted protein structures and experimental structures.

\begin{figure}[!htbp]
	\begin{center}
		\includegraphics[width=1.0\linewidth]{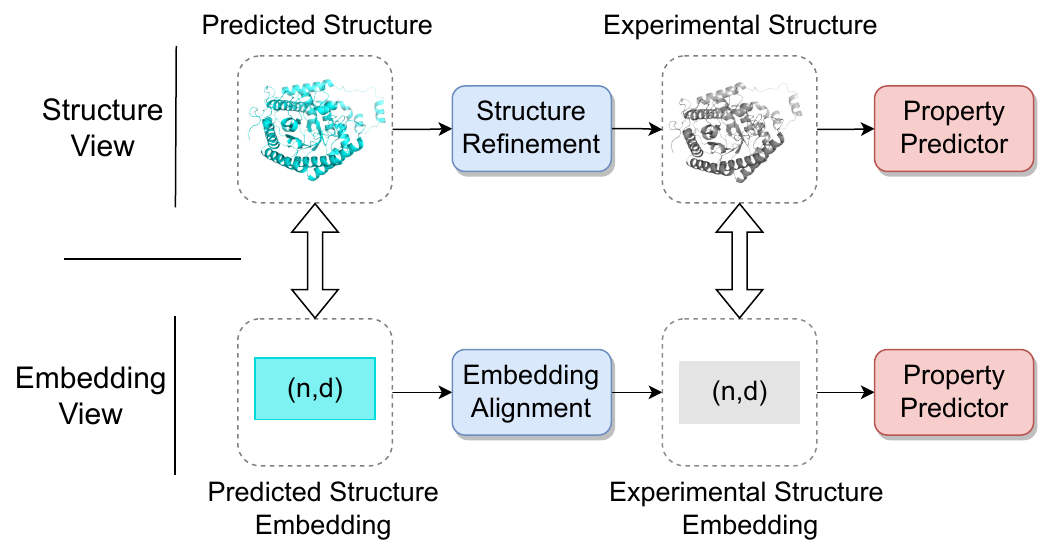}
	\end{center}
	\vspace{-0.5em}
	\caption{An illustration of PGSL-RP3 and two different views for understanding the problem. The alignment (or protein 3D graph structure learning) from the predicted structure to the experimental structure can occur in either the structure space or the embedding space.}
	\label{fig:2}
\end{figure}

We propose two different views for understanding PGSL-RP3 in Fig.\ref{fig:2}, which also correspond to two types of solution ideas. The second one is adopted in our \texttt{SAO} framework. Further preliminaries can be found in Appendix B.

\section{Protein Structure Embedding Alignment Optimization Framework}
\subsection{Problem Analysis and Motivation}
The comparisons between PGSL and GSL motivate us to analyze PGSL from the perspective of embedding. PGSL is similar to GSL in that they both expect a cleaner graph structure to improve performance on downstream tasks. GSL usually focuses on node classification tasks on 2D graphs, where the modifications of the graph structure are generally discrete operations such as the addition and deletion of edges. Different from GSL, PGSL focuses on graph classification tasks on 3D graphs, similar to 3D point clouds, where the modifications of the graph structure are continuous operations, such as the modification of node coordinates. As a result, it is difficult to migrate methods directly from the GSL domain to PSGL. However, the idea of node embedding alignment in unsupervised graph structure learning inspires us to explore PGSL from the embedding view. 

\begin{figure*}[!tbp]
    \begin{center}
        \hspace{-3.5em}
        \subfigure[The \texttt{SAO} Framework]{\includegraphics[width=0.6\linewidth]{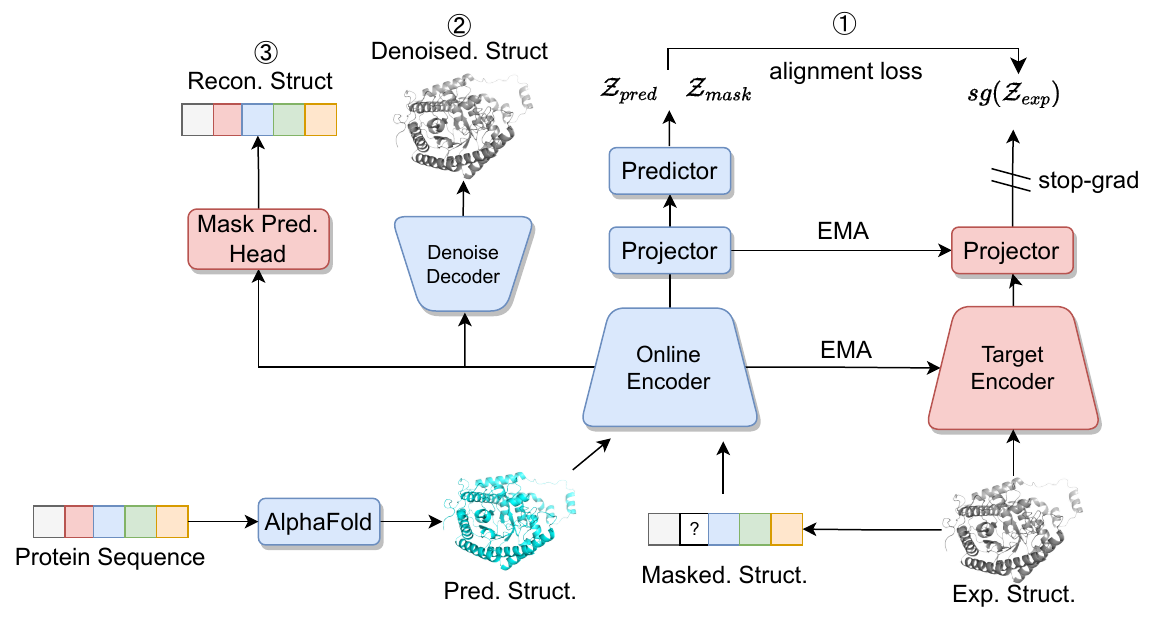}\label{fig:3a}}
        \subfigure[The finetuning and inference scheme]{\includegraphics[width=0.3\linewidth]{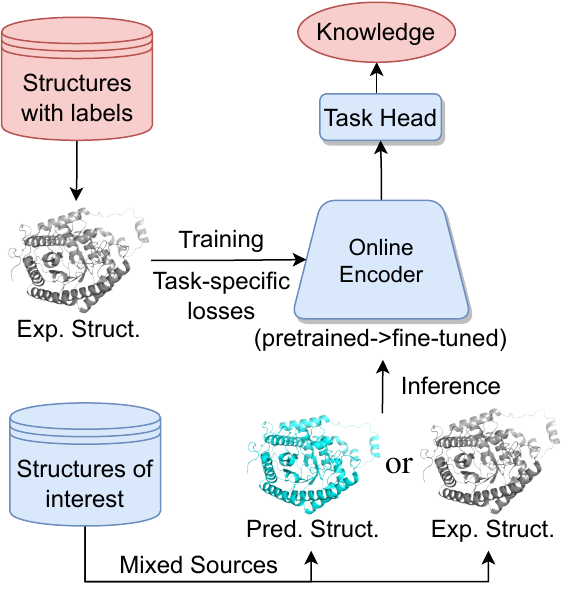}\label{fig:3b}} 
    \end{center}
    \vspace{-1em}
    \caption{The illustration of \texttt{SAO} Framework. (a) We first pretrain the encoder in \texttt{SAO} framework for the ability of directional embedding alignment (b). Then we finetune the encoder on specific downstream tasks. Finally, we can turn structures of interest (predicted or experimental) into new knowledge(predict its various properties) to guide scientific discovery.}
    \label{fig:3}
\end{figure*}

As a result, we propose the \texttt{SAO} framework for alignment in Embedding Space as illustrated in Fig.\ref{fig:3a}. Instead of optimizing the structure directly, our framework used the idea of representation learning to move the embedding of predicted structures toward that of the experimental structures. In practice, we first pretrain the protein encoder based on the \texttt{SAO} framework to give it the ability to migrate representations of predicted structures towards that of experimental structures. Then, by fine-tuning the encoder on specific tasks, as in Fig.\ref{fig:3b}, the model learns to associate representations and labels together. Since the property prediction problem is essentially carried out based on the structure embedding, such an idea can instead improve the model robustness more directly. Our proposed framework will be discussed in the following subsections. 
\subsection{Directional Embedding Alignment}
The motivation of the first module is to help the model achieve \emph{Direction Embedding Alignment}. Noting that there is actually only some natural mapping relationship between the experimental protein structure and its properties, we should learn the correlation between protein structure and properties based on the experimental structure. Therefore, we need to align the embedding of the predicted structure with the embedding of the experimental structure. 

To achieve this, we propose a one-way embedding alignment method, drawing on bootstrap-based contrastive learning methods~\cite{grill2020bootstrap, chen2020exploring}. Our method is based on the student-teacher architecture, where the parameters $\phi$ of the teacher network (a.k.a. Target network) are the exponential moving average of the parameters $\theta$ of the student network (also known as Online network) and only the parameters of the student encoder are trainable. More precisely, given a target decay rate $\lambda$, the following updates are performed after each training iteration,
\begin{equation}
\begin{small}
\phi \leftarrow \lambda \phi+(1-\lambda)\theta.
\label{equ:1}
\end{small}
\end{equation}
The online network is comprised of three stages: an encoder $f_{\theta}$, a projector $g_{\theta}$, and a predictor $q_{\theta}$. The target network and online network have the same architecture, except that there is no predictor. The embedding outputs of both the student encoder and teacher encoder are fed into an embedding projector as shown in Figure \ref{fig:3a}.

Given a set $\mathcal{S}$ of paired protein structures, a protein structure pair sampled uniformly from $\mathcal{S}$, the teacher network takes the experimental structure as input and outputs a representation $z_{\phi}$ of the experimental structure as the prediction target. The student network takes the predicted structure as input. It outputs a representation $z_{\theta}$ of the predicted structure that goes through the predictor $q_{\theta}$ to be aligned with the experimental structure representation. We can $\ell_2$-normalize both $q_{\theta}(z_{\theta})$ and $z_{\phi}$ to $\overline{q_\theta}(z_\theta)\triangleq q_\theta(z_\theta)/\left\|q_\theta(z_\theta)\right\|_2$ and $\overline{z}_\phi\triangleq(z_\phi)/\left\|(z_\phi)\right\|_2$. The loss function is as follow:
\vspace{0.5em}
\begin{equation}
\begin{small}
\mathcal{L}^{\theta,\phi}_{align}\triangleq\left\|\overline{q_\theta}(z_\theta)-\overline{z}_\phi \right\|_2^2=2-2\cdot\frac{\langle q_\theta(z_\theta),z_\phi\rangle}{\left\|q_\theta(z_\theta)\right\|_2\cdot\left\|z_\phi\right\|_2},
\label{equ:2}
\end{small}
\vspace{0.5em}
\end{equation}
where the first part (i.e., $\ell2$-norm loss) is equal to cosine similarity loss.

The loss $\mathcal{L}^{\theta,\phi}_{align}$ is asymmetric to ensure directional embedding alignment, and the network always uses the representation of the predicted structure to predict the representation of the experimental structure. It is worth noting that we only optimize the loss $\mathcal{L}^{\theta,\phi}_{align}$ according to $\theta$, for the target network with $\phi$ we use a gradient-stopping technique. After pretraining, we only keep the online encoder as shown in Figure \ref{fig:3b}.
\subsubsection{Embedding Bootstrap Mechanism}
To give the encoder persistent guidance and avoid over-fitting in the experimental representation, we use the embedding bootstrap mechanism to provide a self-enhanced experimental structure embedding as the learning target. Considering that the embedding is ultimately reflected in the parameters, our core idea is to update the target encoder with an exponential moving average of the learned online encoder as in equation \ref{equ:1} instead of keeping it unchanged.

\subsection{Mask View as Embedding Augmentation}
The feasible solutions in the above framework include collapsed representation (e.g., constant representation across sources is always fully predictive of itself). However, $\mathcal{L}^{\theta,\phi}_{align}$ is not a loss such that \texttt{SAO}'s dynamics is a gradient descent on $\mathcal{L}$ jointly over $\theta, \phi$. There is, therefore, no a \emph{priori} reason why \texttt{SAO}'s parameters would converge to an undesirable minimum of $\mathcal{L}^{\theta,\phi}_{align}$~\cite{grill2020bootstrap}.

Furthermore, to avoid the undesirable equilibria in \texttt{SAO}'s dynamics, we introduce the Mask view as Embedding Augmentation (as shown in Fig.\ref{fig:3a}). We mask the protein sequence of the experimental structure and keep the structure visible to create a mask view. Then we align the embedding of the masked structure toward that of the experimental structure following the same procedure we operate on the predicted structure. It has the following two advantages.

\subsubsection{Increase Variability for Avoiding Collapse}
The technical consideration behind this is to propagate new sources of variability captured by the online projection into the target projection. As shown in ~\cite{grill2020bootstrap}, increasing variability across sources could make these collapsed constant equilibria unstable. The variability between the predicted and experimental structure is limited because, as we mentioned above, the difference between the two is mainly in \textbf{\emph{embedding bias}} when the prediction accuracy is high.

\subsubsection{Embedding Augmentation for Better Alignment}
Another motivation behind introducing the mask view is to increase the ability of the framework to align embedding from other sources to the experimental structural embedding. Due to the small variability between the predicted and experimental structures, the model is more prone to overfitting and loss of generalizability. The predicted structure mainly carries global variability, so we consider proposing augmentation with \textbf{\emph{contextual variability}} to enhance the ability of directional embedding alignment of our framework.

\subsection{Structure Denoising Guidance}
\label{Denoising}
To ensure meaningful representations that can be decoded into cleaner structures and guide the directional embedding alignment, we incorporate a structure denoising task.

Instead of reducing the amino acid molecule to a single $C_{\alpha}$ atom, we pursued a finer-grained modeling, so we considered all the amino acid backbone atoms $(C, N, O, C_{\alpha})$. Considering physical plausibility (the bond lengths and bond angles between backbone atoms are relatively fixed, i.e., the relative positions between backbone atoms can be considered fixed), we modeled the backbone atoms as a frame, such that an amino acid backbone structure is determined by two vector properties: translation ${\mathbf{t}}$ (coordinates of the $C_{\alpha}$ atom) and orientation $\mathcal{O}$ (which determines the final coordinates):
\vspace{0.5em}
\begin{equation}
\begin{small}
\mathcal{P} = \{(\mathbf{t}_i, \mathcal{O}_i)\}_{i=1}^{i=N} \ \ \ \text{and} \ \ \ x^a_i = \mathcal{O}_ix^a_{stand} + \mathbf{t}_i\cdot
\label{equ:3}
\end{small}
\vspace{0.5em}
\end{equation}
where $\mathcal{P}$ is the protein backbone structure, \textit{N} is the length of the protein sequence, $x^a_i$ is the $a$-type backbone atom of residue \textit{i} ($a\in {C, N, O, C_{\alpha}}$). $x^a_{stand}$ is the standard coordinate of the $a$-type amino acid backbone atom when the $C_{\alpha}$ atom is at the origin and under the unit orthogonal group~\cite{jumper2021highly}.

\begin{table*}[!htbp]
\setlength{\tabcolsep}{4pt}
\begin{center}
\resizebox{1.0\textwidth}{!}{
\vspace{-1em}
\scriptsize
\begin{tabular}{@{}clcccccccc@{}}
\toprule
\multirow{2}{*}{Models}       & \multicolumn{1}{c}{\multirow{2}{*}{Inference Setting}} & \multicolumn{2}{c}{EC}      & \multicolumn{2}{c}{GO-MF}                & \multicolumn{2}{c}{GO-CC}   & \multicolumn{2}{c}{GO-BP}   \\ \cmidrule(l){3-10} 
                              & \multicolumn{1}{c}{}                                   & Fmax         & AUPR         & Fmax         & AUPR                      & Fmax         & AUPR         & Fmax         & AUPR         \\ \midrule
\multirow{4}{*}{IPA-Encoder}  & Predicted (TM $\geq$ 0.5)                              & 0.568 \textcolor{black}{\textbf{(+0.156)}} & 0.505 \textcolor{black}{\textbf{(+0.187)}} & 0.492 \textcolor{black}{\textbf{(+0.055)}} & 0.493 \textcolor{black}{\textbf{(+0.071)}} 
                                                                                       & 0.383 \textcolor{black}{\textbf{(+0.061)}} & 0.174 \textcolor{black}{\textbf{(+0.072)}} & 0.311 \textcolor{black}{\textbf{(+0.039)}} & 0.194 \textcolor{black}{\textbf{(+0.027)}} \\
                              & Predicted (pLDDT $\geq$ 70)                            & 0.579 \textcolor{black}{\textbf{(+0.158)}} & 0.521 \textcolor{black}{\textbf{(+0.203)}} & 0.494 \textcolor{black}{\textbf{(+0.056)}} & 0.491 \textcolor{black}{\textbf{(+0.068)}}                     
                                                                                       & 0.362 \textcolor{black}{\textbf{(+0.084)}} & 0.167 \textcolor{black}{\textbf{(+0.082)}} & 0.312 \textcolor{black}{\textbf{(+0.040)}} & 0.198 \textcolor{black}{\textbf{(+0.037)}} \\
                              & Experimental Structure                                 & 0.711 \textcolor{black}{\textbf{(+0.054)}} & 0.707 \textcolor{black}{\textbf{(+0.046)}} & 0.522 \textcolor{black}{\textbf{(+0.025)}} & 0.518 \textcolor{black}{\textbf{(+0.040)}}                    
                                                                                       & 0.433 \textcolor{black}{\textbf{(+0.024)}} & 0.248 \textcolor{black}{\textbf{(+0.015)}} & 0.338 \textcolor{black}{\textbf{(+0.023)}} & 0.218 \textcolor{black}{\textbf{(+0.016)}} \\
                              & The Performance Gap                                   & -0.132 \textcolor{black}{\textbf{(+0.104)}}& -0.186 \textcolor{black}{\textbf{(+0.157)}}& -0.028 \textcolor{black}{\textbf{(+0.031)}}& -0.027 \textcolor{black}{\textbf{(+0.028)}}                   
                                                                                       & -0.071 \textcolor{black}{\textbf{(+0.060)}}& -0.081 \textcolor{black}{\textbf{(+0.067)}}& -0.026 \textcolor{black}{\textbf{(+0.017)}}& -0.02 \textcolor{black}{\textbf{(+0.021)}} \\ \midrule
\multirow{4}{*}{Uni-Mol}      & Predicted (TM $\geq$ 0.5)                              & 0.520 \textcolor{black}{\textbf{(+0.315)}} & 0.475 \textcolor{black}{\textbf{(+0.341)}} & 0.385 \textcolor{black}{\textbf{(+0.195)}} & 0.385 \textcolor{black}{\textbf{(+0.184)}}                    
                                                                                       & 0.333 \textcolor{black}{\textbf{(+0.078)}} & 0.152 \textcolor{black}{\textbf{(+0.083)}} & 0.267 \textcolor{black}{\textbf{(+0.115)}} & 0.149 \textcolor{black}{\textbf{(+0.107)}}       \\
                              & Predicted (pLDDT $\geq$ 70)                            & 0.528 \textcolor{black}{\textbf{(+0.305)}} & 0.486 \textcolor{black}{\textbf{(+0.332)}} & 0.397 \textcolor{black}{\textbf{(+0.187)}} & 0.391 \textcolor{black}{\textbf{(+0.176)}}
                                                                                       & 0.343 \textcolor{black}{\textbf{(+0.068)}} & 0.162 \textcolor{black}{\textbf{(+0.077)}} & 0.267 \textcolor{black}{\textbf{(+0.115)}} & 0.151 \textcolor{black}{\textbf{(+0.108)}}       \\
                              & Experimental Structure                                 & 0.657 \textcolor{black}{\textbf{(+0.155)}}  & 0.643 \textcolor{black}{\textbf{(+0.150)}} & 0.415 \textcolor{black}{\textbf{(+0.156)}} & 0.398 \textcolor{black}{\textbf{(+0.142)}}
                                                                                       & 0.402 \textcolor{black}{\textbf{(+0.030)}} & 0.203 \textcolor{black}{\textbf{(+0.056)}} & 0.314 \textcolor{black}{\textbf{(+0.070)}} & 0.203 \textcolor{black}{\textbf{(+0.049)}}       \\
                              & The Performance Gap                                   & -0.129 \textcolor{black}{\textbf{(+0.150)}}& -0.157 \textcolor{black}{\textbf{(+0.182)}}& -0.018 \textcolor{black}{\textbf{(+0.031)}}& -0.007 \textcolor{black}{\textbf{(+0.034)}}
                                                                                       & -0.059 \textcolor{black}{\textbf{(+0.038)}}& -0.041 \textcolor{black}{\textbf{(+0.021)}}& -0.047 \textcolor{black}{\textbf{(+0.045)}}& -0.052 \textcolor{black}{\textbf{(+0.059)}}      \\ \midrule
\multirow{4}{*}{REINet} & Predicted (TM $\geq$ 0.5)                              & 0.584 \textcolor{black}{\textbf{(+0.147)}} & 0.506 \textcolor{black}{\textbf{(+0.193)}} & 0.519 \textcolor{black}{\textbf{(+0.028)}} & 0.521 \textcolor{black}{\textbf{(+0.033)}}
                                                                                       & 0.337 \textcolor{black}{\textbf{(+0.105)}} & 0.171 \textcolor{black}{\textbf{(+0.054)}} & 0.340 \textcolor{black}{\textbf{(+0.032)}} & 0.227 \textcolor{black}{\textbf{(+0.020)}}       \\
                              & Predicted (pLDDT $\geq$ 70)                            & 0.599 \textcolor{black}{\textbf{(+0.153)}} & 0.546 \textcolor{black}{\textbf{(+0.184)}} & 0.526 \textcolor{black}{\textbf{(+0.030)}} & 0.518 \textcolor{black}{\textbf{(+0.039)}}
                                                                                       & 0.335 \textcolor{black}{\textbf{(+0.104)}} & 0.177 \textcolor{black}{\textbf{(+0.048)}} & 0.340 \textcolor{black}{\textbf{(+0.033)}} & 0.229 \textcolor{black}{\textbf{(+0.024)}}       \\
                              & Experimental Structure                                 & 0.756 \textcolor{black}{\textbf{(+0.038)}} & 0.755 \textcolor{black}{\textbf{(+0.026)}} & 0.535 \textcolor{black}{\textbf{(+0.013)}} & 0.540 \textcolor{black}{\textbf{(+0.011)}}
                                                                                       & 0.427 \textcolor{black}{\textbf{(+0.021)}} & 0.251 \textcolor{black}{(-0.010)}    & 0.360 \textcolor{black}{\textbf{(+0.019)}} & 0.248 \textcolor{black}{\textbf{(+0.007)}}       \\
                              & The Performance Gap                                   & -0.157 \textcolor{black}{\textbf{(+0.115)}}& -0.209 \textcolor{black}{\textbf{(+0.158)}}& -0.009 \textcolor{black}{\textbf{(+0.017)}}& -0.022 \textcolor{black}{\textbf{(+0.028)}}
                                                                                       & -0.092 \textcolor{black}{\textbf{(+0.083)}}& -0.074 \textcolor{black}{\textbf{(+0.058)}}& -0.020 \textcolor{black}{\textbf{(+0.014)}}& -0.019 \textcolor{black}{\textbf{(+0.017)}}      \\ \bottomrule
\end{tabular}}
\end{center}
\vspace{-1em}
\caption{Downstream task performance of three inference settings across three encoders. Results in parentheses (\textcolor{black}{\textbf{bold}}) are performance gain training with \texttt{SAO} framework, the performance gap is calculated by values of Predicted(Plddt $\geq$ 70) - values of Experimental Structure. Results clearly show \texttt{SAO} is effective across tasks and model-agnostic.}
\label{tab:1}
\end{table*}
Given a predicted protein structure $\mathcal{P}$, the online encoder maps it into embedding. A denoise decoder head predicts the direction of conformational change for further optimization. Formally, we can formulate this process as follows:
\vspace{0.5em}
\begin{equation}
\begin{small}
\begin{aligned}
\{(\Delta\mathbf{t}_i, \Delta\mathcal{O}_i)\}_{i=1}^{i=N} = \mathcal{E}(\mathcal{P}), \\
\mathbf{t}_i = \mathbf{t}_i + \Delta\mathbf{t}_i\ \text{and}\ \mathcal{O}_i = \mathcal{O}_i \circ \Delta\mathcal{O}_i\cdot
\label{equ:4}
\end{aligned}
\end{small}
\vspace{0.5em}
\end{equation}
where $\mathcal{E}$ is the equivariant neural network and $\circ$ corresponds to the composition of elements in SO(3) group. we use the MSE loss under the local frame~\cite{jumper2021highly} as the training objective:
\vspace{0.5em}
\begin{equation}
\begin{small}
\mathcal{L}_{mse} = \text{Mean}_{i,j}\sqrt{\Vert T^{-1}_i \circ x_{j} - T_i^{true-1} \circ x^{true}_{j} \Vert^2}\cdot
\label{equ:5}
\end{small}
\vspace{0.5em}
\end{equation}
where $i\in\{1,\cdots, N_{res}\}, j\in\{C,N,O,C_{\alpha}\}$, $T_i$ and 
$T^{true}_i$ corresponds to the predicted and ground truth frame $(\mathcal{O},\mathbf{t})$ of residue i. And $T_i^{-1} \circ x_j = \mathcal{O}^{-1}_i(x_j-t_i)$ converts the coordinates of backbone atoms from the global to local frame. More details can be referred to Appendix. C.
\subsection{Overall Framework}
In this subsection, we first illustrate the training process of \texttt{SAO}, and then briefly introduce our model architecture.
\noindent \textbf{Model training.} In our training process, we first pretrain the protein structure encoder in \texttt{SAO} framework on the corresponding downstream task dataset (i.e., no new sample is added during pretrain). The overall pretraining objective is $L_{SAO} = \gamma_1L_{align} + \gamma_2L_{mlm} + \gamma_3L_{mse}$ as figure \ref{fig:3a} shows. $L_{mlm}$ is the standard mask language modeling loss~\cite{devlin2019bert}, and $\gamma$ is the loss weight (value setups are shown in Appendix. E). Then we fine-tune the encoder on the downstream task using task-specific losses(as shown in fig.\ref{fig:3b}). The algorithmic description and more details, including the mask ratio, are provided in Appendix E.

\noindent \textbf{Frame-Aware 3D Graphformer.}
 Our \texttt{SAO} framework is model-agnostic, allowing for compatibility with various models. Additionally, we propose a simple yet effective encoder named \textbf{REInet} that can co-model the 3D geometry of amino acid coordinates and the 1D protein sequences. As the model architecture is not our main concern, we defer its detailed introduction to the Appendix. D.
\begin{table*}[!htbp]
\renewcommand{\arraystretch}{0.8}
\begin{center}
\vspace{-1em}
\scriptsize
\begin{tabular}{@{}clcccccccc@{}}
\toprule
\multirow{2}{*}{Inference Setting}                                                 & \multicolumn{1}{c}{\multirow{2}{*}{\begin{tabular}[c]{@{}c@{}}Robust Training \\ Framework\end{tabular}}} & \multicolumn{2}{c}{EC}                     & \multicolumn{2}{c}{GO-MF}       & \multicolumn{2}{c}{GO-CC}                        & \multicolumn{2}{c}{GO-BP}       \\ \cmidrule(l){3-10} 
                                                                                   & \multicolumn{1}{c}{}                                           & Fmax           & AUPR                      & Fmax           & AUPR           & Fmax           & AUPR                            & Fmax           & AUPR           \\ \midrule
\multirow{6}{*}{TMscore $\geq$ 0.5}                                                     & Vanilla                                                        & 0.584          & 0.506                     & 0.519          & 0.521          & 0.337          & 0.171                           & 0.340          & 0.227          \\
                                                                                   & TonP                                   & 0.731          & 0.693                     & 0.537          & 0.547          & 0.414          & \textbf{0.266}                  & 0.371          & \textbf{0.263} \\
                                                                                   & Mixed                                                 & 0.667          &0.620 & 0.534          & 0.537          & 0.440          & 0.234                           & 0.365          & 0.257          \\
                                                                                   & RefthenPred                                              & 0.562          & 0.616                     & 0.524          & 0.524          & 0.341          & 0.184                           & 0.345          & 0.231          \\
                                                                                   & RefandPred                                               & 0.667          & 0.623                     & 0.501          & 0.505          & 0.396          & 0.194                           & 0.337          & 0.224          \\
                                                                                   & \textbf{\texttt{SAO}(ours)}                                                            & \textbf{0.731} & \textbf{0.699}            & \textbf{0.547} & \textbf{0.554} & \textbf{0.442} & \multicolumn{1}{l}{{\ul 0.225}} & \textbf{0.372} & {\ul 0.247}    \\ \midrule
\multirow{6}{*}{pLDDT $\geq$ 70}                                                        & Vanilla                                                        & 0.599          & 0.546                     & 0.526          & 0.518          & 0.335          & 0.177                           & 0.340          & 0.229          \\
                                                                                   & TonP                               & 0.750          & 0.716                     & 0.544          & 0.547          & 0.419          & \textbf{0.273}                  & 0.371          & 0.269          \\
                                                                                   & Mixed                                                & 0.675          & 0.635                     & 0.542          & 0.534          & 0.433          & 0.224                           & 0.366          & \textbf{0.265} \\
                                                                                   & RefthenPred                                              & 0.627          & 0.587                     & 0.531          & 0.523          & 0.336          & 0.188                           & 0.347          & 0.235          \\
                                                                                   & RefandPred                                               & 0.680           & 0.644                     & 0.511          & 0.508          & 0.400          & 0.212                           & 0.337          & 0.229          \\
                                                                                   & \textbf{\texttt{SAO}(ours)}                                                            & \textbf{0.752} & \textbf{0.730}            & \textbf{0.556} & \textbf{0.557} & \textbf{0.439} & {\ul 0.225}                     & \textbf{0.373} & {\ul 0.253}    \\ \midrule
\multirow{6}{*}{\begin{tabular}[c]{@{}c@{}}Experimental \\ Structure\end{tabular}} & Vanilla                                                        & 0.746          & 0.745                     & 0.535          & 0.540          & 0.427          & 0.251                           & 0.360          & 0.248          \\
                                                                                   & TonP                                   & 0.543          & 0.430                     & 0.388          & 0.357          & 0.334          & 0.165                           & 0.279          & 0.173          \\
                                                                                   & Mixed                                                 & 0.676          & 0.575                     & 0.520          & 0.510          & 0.432          & 0.215                           & 0.364          & 0.247          \\
                                                                                   & RefthenPred                                              & 0.746          & 0.745                     & 0.535          & 0.54           & 0.427          & 0.251                           & 0.36           & 0.248          \\
                                                                                   & RefandPred                                               & 0.742          & 0.730                     & 0.512          & 0.506          & 0.425          & 0.221                           & 0.345          & 0.227          \\
                                                                                   & \textbf{\texttt{SAO}(ours)}                                                            & \textbf{0.784} & \textbf{0.771}            & \textbf{0.548} & \textbf{0.551} & \textbf{0.448} & \textbf{0.241}                  & \textbf{0.379} & \textbf{0.255} \\ \bottomrule
\end{tabular}
\end{center}
\vspace{-1em}
\caption{Performance comparison across different robust training framework baselines. Our \texttt{SAO} surpasses most baselines across different tasks and inference settings, which shows its effectiveness and advantage of using paired protein structure data. The best and second results are marked by \textbf{bold} and \ul{underline}. }
\label{tab:2}
\end{table*}
\section{Experiments}
In this section, we first introduce the experimental setup for four standard protein property prediction tasks, including Enzyme Commission number prediction, and Gene Ontology term prediction following ~\cite{gligorijevic2021structure}, model architectures, and robust training framework baselines. We then conduct empirical experiments to demonstrate the effectiveness of the proposed framework \texttt{SAO}. We aim to answer five research questions as follows: 
\textbf{Q1}: Does the protein structure embedding bias problem generally exist in various protein property prediction tasks across different predictors?
\textbf{Q2}: Are our proposed framework \texttt{SAO} model-agnostic? 
\textbf{Q3}: How effective is \texttt{SAO} for PGSL-RPA?
\textbf{Q4}: How do key framework components impact the performance of \texttt{SAO}?
\textbf{Q5}: How robust is \texttt{SAO} to less accurate or non-AlphaFold predicted protein structures?
\subsection{Experimental Setups}
\subsubsection{Downstream tasks for evaluation}
We adopt four tasks proposed in ~\cite{gligorijevic2021structure} as downstream tasks for evaluation. \textbf{\emph{Enzyme Commission (EC) number prediction}} aims to forecast the EC numbers of various proteins, which describe their catalysis of biological activities in a tree structure. It's a multi-label classification task with 538 categories. \textbf{\emph{Gene Ontology(GO) term prediction}} aims to annotate a protein with GO terms that describe the Molecular Function (MF) of gene products, the Biological Processes (BP) in which those actions occur, and the Cellular Component (CC) where they are present. Thus, GO term prediction is actually composed of three different sub-tasks: \textbf{\emph{GO-MF, GO-BP, GO-CC}}. And two metrics are employed, including the maximum F-score (Fmax) and AUPR~\cite{zhang2022protein}.
\subsubsection{Inference Setting}
To evaluate \texttt{SAO} in realistic scenarios, we choose three inference settings: 1) inference with the experimental structure; 2) inference with predicted protein structure with TMScore $\geq$ 0.5 3) inference with predicted protein structure with pLDDT $\geq$ 70. TMscore and pLDDT are commonly used criteria for selecting accurately predicted protein structures for downstream applications.
\subsubsection{Model architectures}
We explore the PGSL-RAP problem in various protein structure encoder architectures, including graph neural network-based GearNet\_Edge~\cite{zhang2022protein}, graphormer-based Uni-Mol~\cite{zhou2022uni}, equivariant neural network IPA~\cite{jumper2021highly} and 3D Graphormer REInet. Implementation and training details can be referred to Appendix. D and E. For simplicity, we use REInet as the encoder in subsequent experiments unless otherwise specified. 
\subsubsection{Baselines and Training}
We mainly compare \texttt{SAO} with two categories of methods, including three directly training-based methods: 1. vanilla method of training on experimental protein structures of downstream tasks(\emph{Vanilla}); 2. training on predicted protein structure(\emph{TonP}); 3. training on the mixture of predicted and experimental protein structure for direct adaptation(\emph{Mixed}). And two protein structure refinement(e.g., improving the similarity of the predicted and experimental structures directly in Euclidean space) based methods are further included: 1. \emph{RefthenPred} which refine the protein structure before predicting its property; 2. \emph{RefandPredict} which refine the protein structure while predicting its property. The \emph{RefthenPred} is like unsupervised GSL, which obtains clean graph structure without supervision signals; we use the SOTA protein structure refinement method ATOMRefine~\cite{wu2022atomic} for refinement. And the \emph{RefandPred} is like supervised GSL, which refines the graph structure under supervision. We add one layer of simplified Structure Module~\cite{jumper2021highly} after the structure encoder to refine the protein structure while predicting protein properties. 

Following ~\cite{gligorijevic2021structure}, we use the multi-cutoff split methods for EC and GO tasks to guarantee the test set contains only PDB chains with sequence identity less than 95\% to the training set. We pretrain encoders under \texttt{SAO} for 400 epochs on structures of corresponding downstream tasks and then fine-tune them on downstream tasks for specific epochs (EC: 100 epochs, GO-CC: 45 epochs, GO-MF and GO-BP: 100 epochs). Warmup and the exponential learning rate decay schedule are used with a start learning rate of 0.0, a max learning rate of 1e-4, and a decay factor of 0.99. For other experimental details, interested readers can refer to the Appendix.
\begin{figure}[!tbp]
    \begin{center}
    \hspace{-1em}
        \subfigure[EC]{\includegraphics[width=0.5\linewidth]{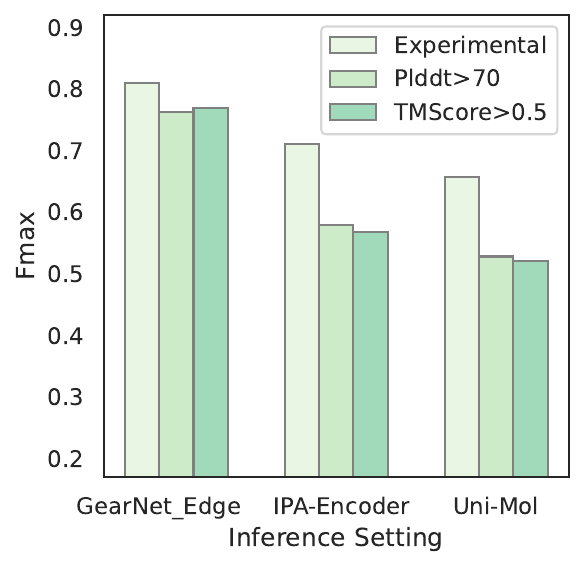}\label{fig:4a}}
        \subfigure[GO-CC]{\includegraphics[width=0.5\linewidth]{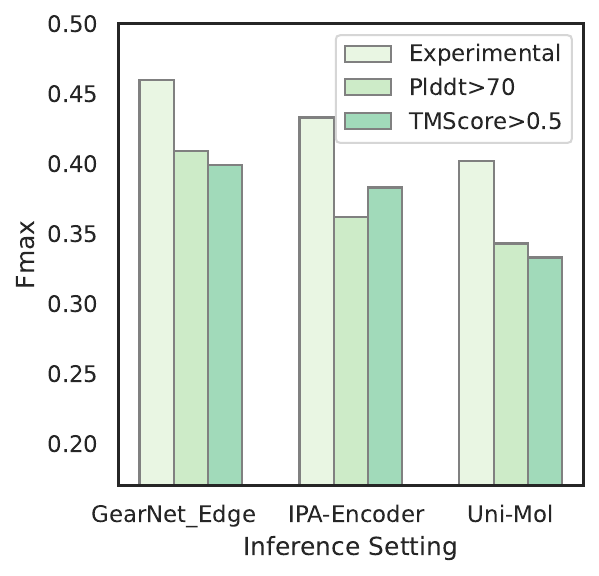}\label{fig:4b}}
    \hspace{-1.5em}
    \end{center}
    \vspace{-1em}
    \caption{Fmax on EC and GO-CC tasks across different encoder architectures in different inference settings. Representative encoders suffer from performance degradation on predicted structures.}
    \label{fig:4}
    \vspace{-1.5em}
\end{figure}
\subsection{Empirical studies (Q1)}
Fig.\ref{fig:4} shows how well different methods predict certain properties of proteins in various tasks. The methods are trained on experimental protein structures, following ~\cite{zhang2022protein}. Then, they are tested in three different inference settings, including two settings with predicted protein structures. As can be observed, all three models perform worse when making predictions with predicted protein structures. This suggests that there is a general issue of protein structure embedding bias, which affects protein property prediction tasks across different prediction methods. Interested readers can find numerical results on more tasks in Appendix. F. 
\subsection{Performance of SAO across different encoders (Q2)}
Table \ref{tab:1} shows the performance of \texttt{SAO} framework across different protein structure encoders. It indicates that \texttt{SAO} framework is model-agnostic and effective in improving the prediction accuracy of both experimental and predicted structures. With \texttt{SAO}, encoders can learn to align the embeddings of predicted structures to that of experimental structures and output meaningful embeddings. It's worth noting that the property prediction performance with predicted protein structures surpasses that with experimental protein structures in some settings, which shows the value of the idea of directional embedding alignment.

\subsection{Performance Comparison (Q3)}
Table \ref{tab:2} reports the classification performance of our framework and other baselines in experimental and predicted protein structure inference scenarios. We can observe that our proposed \texttt{SAO} outperforms all baselines on 8 out of 8 metrics ( 2 for each property prediction task). This superior performance benefits from the novel idea of guiding PGSL in embedding space with a self-enhanced learning target by directional embedding alignment. 

We make other observations as follows. Firstly, the performances of training predictors on experimental or predicted protein structures are both notably biased. It is even worse when training on predicted protein structures. This observation provides another evidence of protein structure embedding bias. Secondly, \texttt{SAO} can surpass all three directly training-based baselines, indicating \texttt{SAO} can make better use of paired experimental-predicted data. Thirdly, compared to refinement-based baselines (whether supervised or not), our unsupervised embedding-based methods also achieve better results, which shows our effectiveness.

\begin{table}[!htbp]
\begin{center}
\resizebox{1.0\columnwidth}{!}{
\begin{tabular}{@{}clcc@{}}
\toprule
\multirow{2}{*}{Inference Setting}      & \multicolumn{1}{c}{\multirow{2}{*}{Robust Training Framework}} & \multicolumn{2}{c}{GO-CC} \\ \cmidrule(l){3-4} 
                                        & \multicolumn{1}{c}{}                                           & Fmax        & AUPR        \\ \midrule
\multirow{6}{*}{TMscore $\geq$ 0.5}          & Vanilla                                                        & 0.337       & 0.171       \\
                                        & SAO                                                            & \textbf{0.442}       & \textbf{0.226}       \\
                                        \cmidrule(l){2-4}
                                        & w/o structure denoising guidance                               & 0.429       & 0.206       \\
                                        & w/o mask view augumentation                                    & 0.388       & 0.201       \\
                                        & w/o embedding alignment                                        & 0.389       & 0.203       \\
                                        & w/o mask language modeling                                     & 0.422       & 0.209       \\ \midrule
\multirow{6}{*}{\begin{tabular}[c]{@{}c@{}}Experimental \\ Structure\end{tabular}} & Vanilla                                                        & 0.427       & 0.251       \\
                                        & SAO                                                            & \textbf{0.449}       & \textbf{0.252}       \\
                                        \cmidrule(l){2-4}
                                        & w/o structure denoising guidance                               & 0.432       & 0.218       \\
                                        & w/o mask view augumentation                                    & 0.412       & 0.240        \\
                                        & w/o embedding alignment                                        & 0.438       & 0.234       \\
                                        & w/o mask language modeling                                     & 0.419       & 0.238       \\ \bottomrule
\end{tabular}}
\end{center}
\vspace{-1em}
\caption{Ablation study for designed components in two inference scenarios. The best metrics are marked by \textbf{bold}.}
\label{tab:3}
\end{table}
\subsection{Ablation Study (Q4)}
To study the importance of every component in \texttt{SAO}, we perform an ablation study 
on every loss term. As shown in Table \ref{tab:3}, without structure denoising guidance, the classification performance decrease by 2\% on average, indicating this component helps improve the quality of learned embedding and the alignment from predicted embeddings to experimental embeddings. Without embedding alignment, we can find obvious drops in classification performance, especially when performing inference on predicted protein structures. It shows the effectiveness of our directional embedding alignment mechanisms. We can further notice a clear decline without the mask view augmentation component, indicating its effectiveness in avoiding embedding collapse and improving directional alignment as embedding augmentation. Finally, it is unsurprising that mask language modeling tasks can improve performance on experimental protein structures. Due to the page limit, more ablation studies, including task weights, can be referred to the Appendix F.
\begin{figure}[!htbp]
	\begin{center}
		\includegraphics[width=1\linewidth]{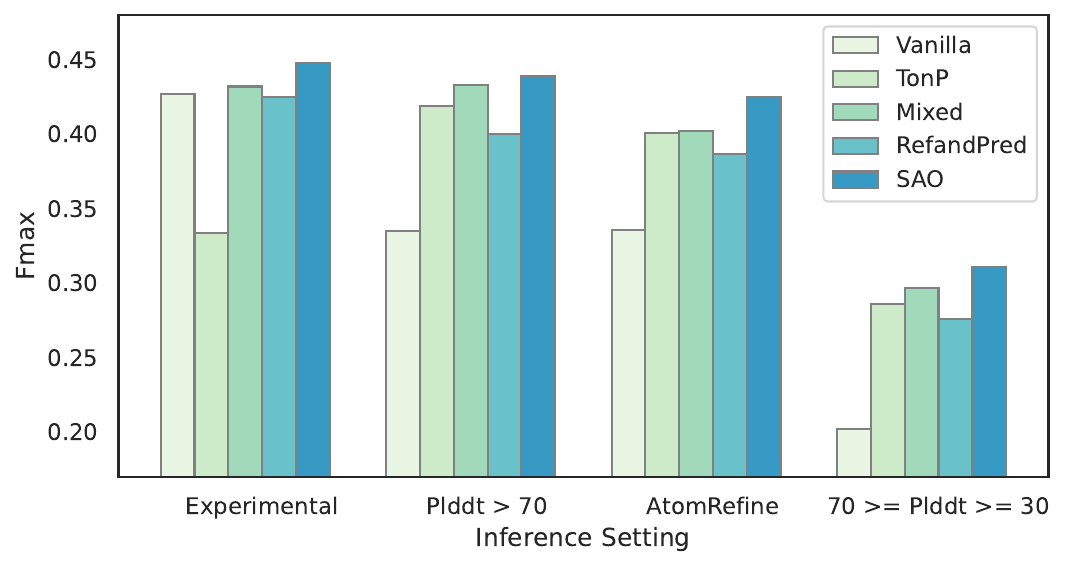}
	\end{center}
        \vspace{-1em}
	\caption{Fmax on GO-CC task in different inference settings. AtomRefine refers to inferencing with ATOMRefine-predicted protein structures. Encoders trained under \texttt{SAO} framework can better handle less accurately predicted protein structures and non-AlphaFold predicted structures.}
	\vspace{-1em}
	\label{fig:5}
\end{figure}
\subsection{More results on predicted structure (Q5)}
Figure \ref{fig:5} shows the performance under less accurate predicted protein structure and non-AlphaFold predicted structure. The results indicate that \texttt{SAO} can make the encoder more robust to less accurate predicted protein structures and generalize well to non-AlphaFold predicted structures. 

We attribute its comparative performance to our design of directional embedding alignment. Directly training-based baselines may struggle to generalize to other inference settings, although have comparative performances in accurate AF-predicted structures.
\section{Conclusion}
This paper investigates 3D protein graph structure learning and designs a novel framework, SAO, which is capable of leveraging pair data to perform directional embedding alignment. To learn the optimal alignment, we employ bootstrap-based contrastive learning to maximize the agreement between predicted structure embeddings and self-enhanced learning targets. Extensive experiments demonstrate the superiority of \texttt{SAO}.

\noindent \textbf{Social Impact and Limitations} This paper identifies an underlying problem of current common practices(i.e., using predicted protein structures from tools like AlphaFold2 as alternatives when experimental protein structures are missing). It might help the community to rethink the usage of massive AlphaFold2 predicted structures. Our study helps improve the property prediction on predicted protein structures. It contributes to the global efforts to transform large amounts of predicted structures into \emph{high-quality} and valuable knowledge for human well-being. Limitations still exist, including insufficient exploration of Protein 3D Graph Structure Learning on non-classification tasks and limited sources of predicted protein structures.

\bibliography{aaai23}

\end{document}


\maketitle
\section{Appendix}

Proteins are involved in various important life processes, such as immune response and DNA replication, so understanding proteins is important for deciphering the mystery of life and treating various diseases.~\cite{jumper2021highly}
\subsection{A. More Related Work}
\noindent \textbf{Protein Structure Refinement.}
The Protein Structure Refinement is to increase the accuracy of the predicted protein structure or known as \emph{Decoy}~\cite{hiranuma2021improved}. In the past, PSR methods were mainly based on molecular dynamics simulations as well as conformational sampling~\cite{adiyaman2019methods}. Subsequent deep learning-based methods attempt to bootstrap traditional optimization tools such as Rosetta by predicting a better protein distance map.~\cite{roney2022state, hiranuma2021improved,jing2020learning}. Following Alphafold2, the field shifted to direct modeling in 3D, using equivariant neural networks to directly obtain atomic-level optimization results~\cite{wu2022atomic}.

\noindent \textbf{Protein Model Quality Assessment.}
Some work uses neural networks to predict the accuracy of Decoy (which is a twin problem of PSR called Model Quality Assessment, MQA) as a way to guide conformational sampling toward the lowest point of folding energy~\cite{hiranuma2021improved}. Other works model at the pseudo-3D level, starting from the scalar feature Distance Map of the protein, optimizing the prediction of the Distance Map by a neural network such as GNN, and then using tools such as Rosetta to perform conformational sampling based on the optimized Distance Map~\cite{jing2021fast}.

\noindent \textbf{DDPM for Protein Structure.}
Denoising Diffusion Probability Model (DDPM)~\cite{sohl2015deep,ho2020denoising}, a recently developed generative model, has been applied to protein 3D structure generation. Its forward diffusion process inspired \emph{RefineDiff}. Unusually, the first pieces of work are the direct generation of 3D structures(not pseudo-3D features)~\cite{anand2022protein}, and until now, it's been the 3D generation that dominated. Subsequent work has seen further developments in application areas, forward diffusion processes~\cite{luo2022antigen}, sampling~\cite{trippe2022diffusion}, conditional generation~\cite{ingraham2022illuminating}, network structures~\cite{watson2022broadly}, etc. Efficient DDPM based on pseudo-3D features (like distance maps~\cite{lee2022proteinsgm}, and torsion angles~\cite{wu2022protein}) have also emerged during this period. Recently, RFDiffusion~\cite{watson2022broadly} reveals the connection between denoising models and structure prediction models, and Chroma~\cite{ingraham2022illuminating} significantly developed the protein conditional generation model.
\begin{figure*}[!htbp]
    \centering
    \includegraphics[width=\linewidth]{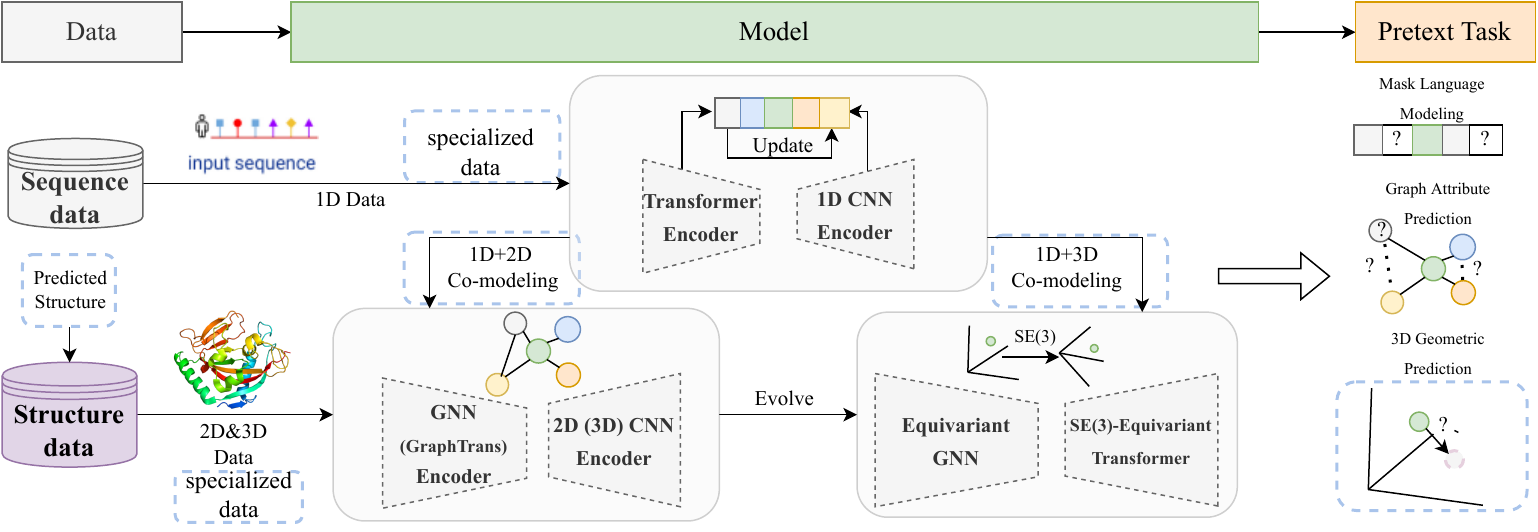}
    \vspace{-1em}
    \caption{A unified framework for protein pretraining. The part framed by the blue line is the direction worth noting. 1D means protein sequence, 2D means 2D protein graph which is based on KNN, 3D means modeling in 3D space or \textbf{3D geometric aware}}
    \vspace{-1em}
    \label{fig:1}
\end{figure*}
\noindent \textbf{Protein Pretraining.}
With the development of protein representation models, pretraining was naturally introduced into the field. The first great successes were protein sequence pretraining. The Evolutionary Scale Modeling (ESM) family of models has achieved remarkable success in difficult downstream tasks such as mutation prediction~\cite{meier2021language} and protein structure prediction models~\cite{hsu2022learning}. With the development of structure prediction tools as well as GNN pretraining, structure pretraining has taken off, and several recent works have advanced the development of pseudo-3D structure pretraining. They typically downscale 3D structures into scalar features on 2D GNNs and migrate the pretext task of 2D GNN Pretraining~\cite{zhang2022protein,hermosilla2022contrastive,chen2022structure}. Thanks to the development of geometric deep learning, end-to-end 3D pretraining started a limited exploration~\cite{mansoor2021toward}, facing the difficulty of insufficient 3D pretraining tasks. The 3D molecular pretraining model Uni-Mol~\cite{zhou2022uni} has shown powerful representation capabilities, and we believe that protein 3D pretraining is an equally promising direction.

\noindent \textbf{Graph Structure Learning}
The contributions of existing work in GSL can essentially be summarized as supervised Graph Structure Learning~\cite{GSL} and unsupervised Graph Structure Learning~\cite{UGSL}. We mainly introduce unsupervised Graph Structure Learning and refer readers to a full survey~\cite{zhu2022survey}.

Insufficient downstream supervision probably leads to the under-supervision problem and may introduce bias to graph structure learning~\cite{UGSL, HE-GSL}. To address this problem, based on the feature-label consistency hypothesis, SLAPS~\cite{fatemi2021slaps} employs the pretext task of feature reconstruction to provide more self-supervision. The graph structure is a core element of GNNs, which makes GSL play a very important role in various graph-related applications. However, most previous work has focused on applying GSL to improve the performance of downstream tasks. Still, little effort has been made to explore “how to optimize the learning of graph structure,” i.e., designing supervision signals for optimizing the structure learning. For example, StructSSL~\cite{StructGSL} presents a probabilistic semi-supervised framework based on sparse GSL, which learns a sparse weighted graph from the unlabeled high-dimensional data and a small amount of labeled data.
\begin{table*}[!htbp]
\setlength{\tabcolsep}{4pt}
\begin{center}
\resizebox{1.0\textwidth}{!}{
\vspace{-1em}
\scriptsize
\begin{tabular}{@{}llllllllllll@{}}
\toprule
Dataset & Max\_lr & Base\_lr & Warmup\_step & start\_decay & decay\_evey\_n\_step & batch\_size & $\gamma_1^{pred}$ & $\gamma_1^{mask}$ & $\gamma_2$ & $\gamma_3$ & Epochs \\ \midrule
EC      & 1e-4    & 0.0      & 3600         & 36000        & 180                  & 128         & 1.0               & 0.1               & 0.5        & 1.0        & 400    \\
GO-BP   & 1e-4    & 0.0      & 3600         & 36000        & 180                  & 128         & 1.0               & 0.1               & 0.5        & 1.0        & 400    \\
GO-MF   & 1e-4    & 0.0      & 3600         & 36000        & 180                  & 128         & 1.0               & 0.1               & 0.5        & 1.0        & 400    \\
GO-CC   & 1e-4    & 0.0      & 3600         & 36000        & 180                  & 128         & 1.0               & 1.0               & 0.5        & 0.1        & 400    \\ \bottomrule
\end{tabular}
}
\end{center}
\vspace{-1em}
\caption{Hyper-parameter setting for pretraining on every task.}
\label{tab:5}
\end{table*}
\begin{table*}[!htbp]
\setlength{\tabcolsep}{4pt}
\begin{center}
\resizebox{0.8\textwidth}{!}{
\vspace{-1em}
\scriptsize
\begin{tabular}{@{}cccccccc@{}}
\toprule
\multicolumn{1}{l}{Dataset} & \multicolumn{1}{l}{Max\_lr} & \multicolumn{1}{l}{Base\_lr} & \multicolumn{1}{l}{Warmup\_step} & \multicolumn{1}{l}{start\_decay} & \multicolumn{1}{l}{decay\_evey\_n\_step} & \multicolumn{1}{l}{batch\_size} & \multicolumn{1}{l}{Epochs} \\ \midrule
EC                          & 1e-4                        & 0.0                          & 2500                             & 25000                            & 200                                      & 32                              & 100                        \\
GO-BP                       & 1e-4                        & 0.0                          & 3600                             & 36000                            & 240                                      & 32                              & 100                        \\
GO-MF                       & 1e-4                        & 0.0                          & 3600                             & 36000                            & 240                                      & 32                              & 100                        \\
GO-CC                       & 1e-4                        & 0.0                          & 3600                             & 10800                            & 360                                      & 32                              & 50                         \\ \bottomrule
\end{tabular}
}
\end{center}
\vspace{-1em}
\caption{Hyper-parameter setting for finetuning on every task.}
\label{tab:6}
\end{table*}

\subsection{B. More Preliminaries}
\subsubsection{Protein Pretraining}
Protein Pretraining usually consists of three elements: data $\mathcal{D}$, representation model $f_{\theta}(\cdot)$, and pretext tasks as losses

$\{\mathcal{L}_{pre}^{\uppercase\expandafter{\romannumeral1}}(\theta, \mathcal{D}), \mathcal{L}_{pre}^{\uppercase\expandafter{\romannumeral2}}(\theta, \mathcal{D}), \cdots, \mathcal{L}_{pre}^{\mathcal{T}}(\theta, \mathcal{D})\}$ (See Fig.\ref{fig:1}). Each task corresponds to a specific projection head $\{g_{T}(\cdot)\}_{T=1}^{\mathcal{T}}$. Protein pretraining is generally performed in two steps: (1) Pretraining the representation model $f_\theta(\cdot)$ with (a) pretext task(s) in pretraining dataset $D_{pre}$; and (2) Fine-tuning the pretrained representation model $f_{\theta_{pre}}(\cdot)$ with a prediction head $g_{downstream}(\cdot)$ under the supervision of a specific downstream task $\mathcal{L}_{task}(\theta, \mathcal{D}_{task})$. The whole process can be formulated as
\setlength\abovedisplayskip{0.3em}
\setlength\belowdisplayskip{0.3em}
\begin{equation}
\begin{small}
\begin{aligned}
\theta^{*} = & \arg \min _{\theta} \mathcal{L}_{task}(\theta_{pre}, \mathcal{D}_{task}), \\ \text{s.t.} \text{ } \text{ } \theta_{pre} = &\mathop{\arg\min}_{\theta} \sum_{k=1}^K \lambda_k \mathcal{L}_{pre}^{(k)}(\theta, \mathcal{D})\cdot
\label{equ:1}
\end{aligned}
\end{small}
\end{equation}
where $\{\lambda_k\}_{k=1}^K$ are task weight hyperparameters. In particular, if we set $K = 1$ and $\mathcal{L}_{pre}^{(1)}(\theta, \mathcal{D}) = \mathcal{L}_{task}(\theta, \mathcal{D}_{task})$, it is equivalent to train from scratch on a downstream task like many previous works on protein representation, which can be considered a special case of our framework.
\subsubsection{Equivariant and Invariant}
The \emph{scalar} and \emph{vector} attributes we mentioned before should be strictly defined as \emph{invariant} or \emph{equivariant} attribute under rotation or translation of the protein. Formally, $f: \mathbb{R}^3 \rightarrow \mathbb{R}^S$ is an \emph{invariant} function \textit{w.r.t} SE(3) group, if for any rotation and translation transformation in the group, which is represented by  $\mathbf{\mathrm{R}}$ as orthogonal matrices and $\mathbf{\mathrm{t}} \in\mathbb{R}^3$   respectively, $f(\mathbf{\mathrm{R}} \mathbf{x} + \mathbf{\mathrm{t}}) = f(\mathbf{x})$. Attributes generated by \emph{invariant} function are \emph{invariant} and \emph{scalar}. If $f: \mathbb{R}^3 \rightarrow \mathbb{R}^3$ is an \emph{equivariant} \textit{w.r.t.} SE(3) group, then $f(\mathbf{\mathrm{R}} \mathbf{x} + \mathbf{\mathrm{t}}) = \mathbf{\mathrm{R}} f(\mathbf{x}) + \mathbf{\mathrm{t}}$. Attributes generated by \emph{equivariant} function are \emph{equivariant} and \emph{vector}. $f(\cdot)$ can be neural networks and definitions still apply.

\subsubsection{Structure View vs Embedding View}
In the structure view, the predicted structure is explicitly aligned to the experimental structure in Euclidean space, which usually requires a structural learning module to directly optimize the 3D protein structure. The optimized predicted structure is closer to the experimental structure, on the basis of which property predictions are then made to obtain more accurate and reliable property predictions that are closer to the experimental structure. However, direct optimization of the structure is very difficult. It requires a more complex network structure, such as an equivariant neural network. The AlphaFold2 prediction is already relatively accurate, and there is still no method to improve its prediction results by a considerable margin. And there is a risk of introducing new structure distributional bias by using a structure learner. That's why we propose the embedding view. In the embedding view, the predicted structure is implicitly aligned to the experimental structure in the embedding space, thus solving the problem of structural distribution bias of the predicted structure more directly. This view avoids the difficult structure refinement problem and addresses the essence of the problem: protein structures are eventually mapped to the embedding space, and the structural distribution bias will eventually become the embedding bias, so achieving embedding alignment optimization is more straightforward.

\subsection{C. Structure Denoise Decoder}
After obtaining the structure embedding as sequential embedding and pair-wise embedding, the representations are fed into corresponding MLPs as 3D prediction heads to predict $SO(3)$ vector and translation vector $\mathbf{t}_{local}$ in the local frame. We obtain $\Delta\mathcal{O}$ by converting $SO(3)$ vector to rotation matrix and calculate $\Delta\mathbf{t}$ in the global frame by $\Delta\mathbf{t}=\mathcal{O}_{cur}\mathbf{t}_{local}$. The frame-based update method ensures that the model is \emph{equivariant} to global translation and rotation. 
\begin{algorithm}[!htbp]
\caption{Algorithm for the Structure Embedding Alignment Optimization framework}
\label{alg:algorithm}
\textbf{Input}: Experimental structures $S_{exp}$ and corresponding predicted structures $S_{pred}$; \# Epoch: E\\
\textbf{Parameter}: Online encoder $f_\theta$, online projector $g_\theta$, predictor $q_\psi$, target network (encoder + projector) $g_\phi$, denoise head $h_{denoise}$, mask prediction head $h_{mlm}$\\
\textbf{Output}: Parameters $\theta$ of $f_\theta$
\begin{algorithmic}[1] 
\STATE Randomly initialize the parameters of all components except the target network. Copy the parameters of the online network to the target network.
\STATE Stop gradient of the target
\FOR{$epoch \in \{0, 1, ..., E-1\}$}
\STATE Select a batch of paired predicted and experimental protein structures and create mask views of experimental structures to form $P = \{S_{exp}, S_{pred}, S_{mask}\}$.
\STATE Compute the representations \\$E_{pred}, E_{mask} = f_\theta(S_{pred}, S_{mask}$, the predictions \\$Z_{pred}, Z_{mask} = q_{\psi}(g_\theta(E_{pred}, E_{mask}))$, \\the target $Z_{exp} = g_{\phi}(S_{exp})$.
\STATE Compute the reconstructed structures and denoised structures: \\$S_{recon}, S_{denoise} = h_{mlm}(E_{mask}), h_{denoise}(E_{pred})$.
\STATE Calculate the total loss $\mathcal{L}_{total}$ and update online model parameters.
\STATE Perform EMA update to target network: \\$\phi \leftarrow \lambda \phi+(1-\lambda)\theta$.
\ENDFOR
\STATE \textbf{return} Online encoder $f_\theta$.
\end{algorithmic}
\end{algorithm}
\subsection{D. Frame-Aware 3D Graphormer}
Our \texttt{SAO} framework is model-agnostic, allowing for compatibility with various models. Additionally, we propose a simple yet effective encoder named \textbf{REInet} that can co-model the 3D geometry of amino acid coordinates and the 1D protein sequences. In REInet, we model protein structures as frames in Section 4 to capture information at a finer granularity. Further, to capture long-range and spatial pair-wise dependencies in proteins, we adopt the Graphormer architecture~\cite{ying2021do} with additional point-wise spatial attention~\cite{jumper2021highly} and pair-type aware 3D spatial positional encoding~\cite{shuaibi2021rotation,zhou2022uni}. 

First, we employ Multiple Layer Perceptrons(MLPs) to generate sequential and pairwise embeddings for protein structure decoys. The sequential embedding MLPs map residue features containing information on the amino acid type, torsion angle, and 3D coordinates of amino acid backbone atoms to embedding vectors $\{m_i\}_{i=1}^{N_{res}}$. And the pairwise embedding MLPs convert the geometric distance map and pair-type aware 3D spatial positional encoding~\cite{ying2021do} into feature vectors $\{z_{ij}\}_{i,j=1}^{N_{res}}$.
\begin{table*}[!htbp]
\begin{center}
\vspace{-2em}
\scriptsize
\begin{tabular}{@{}cccccccccc@{}}
\toprule
\multirow{2}{*}{Models}       & \multirow{2}{*}{Inference Setting} & \multicolumn{2}{c}{EC}                     & \multicolumn{2}{c}{GO-MF}       & \multicolumn{2}{c}{GO-CC}       & \multicolumn{2}{c}{GO-BP}       \\ \cmidrule(l){3-10} 
                              &                                    & Fmax                      & AUPR           & Fmax           & AUPR           & Fmax           & AUPR           & Fmax           & AUPR           \\ \midrule
\multirow{3}{*}{GearNet\_Edge} & Predicted (TMscore $\geq$ 0.5)               & 0.769                     & 0.774          & 0.599          & 0.526          & 0.399          & 0.262          & 0.415          & 0.248          \\
                              & Predicted (pLDDT $\geq$ 70)             & 0.763                     & 0.764          & 0.606          & 0.525          & 0.409          & 0.262          & 0.414          & 0.245          \\
                              & Experimental Structure             & \textbf{0.810}            & \textbf{0.835} & \textbf{0.612} & \textbf{0.570} & \textbf{0.460} & \textbf{0.303} & \textbf{0.420} & \textbf{0.251} \\ \midrule
\multirow{3}{*}{Uni-Mol}      & Predicted (TMscore $\geq$ 0.5)               & \multicolumn{1}{l}{0.520} & 0.475          & 0.385          & 0.385          & 0.333          & 0.152          & 0.267          & 0.149          \\
                              & Predicted (pLDDT $\geq$ 70)             & 0.528                     & 0.486          & 0.397          & 0.391          & 0.343          & 0.162          & 0.267          & 0.151          \\
                              & Experimental Structure             & \textbf{0.657}            & \textbf{0.643} & \textbf{0.415} & \textbf{0.398} & \textbf{0.402} & \textbf{0.203} & \textbf{0.314} & \textbf{0.203} \\ \midrule
\multirow{3}{*}{IPA-Encoder}  & Predicted (TMscore $\geq$ 0.5)               & 0.568                     & 0.505          & 0.492          & 0.493          & 0.383          & 0.174          & 0.311          & 0.194          \\
                              & Predicted (pLDDT $\geq$ 70)             & 0.579                     & 0.521          & 0.494          & 0.491          & 0.362          & 0.167          & 0.312          & 0.198          \\
                              & Experimental Structure             & \textbf{0.711}            & \textbf{0.707} & \textbf{0.522} & \textbf{0.518} & \textbf{0.433} & \textbf{0.248} & \textbf{0.338} & \textbf{0.218} \\ \bottomrule
\end{tabular} \vspace{-1em}
\end{center}
\caption{Performance of three representative protein structure encoders(trained on experimental structures) on four property prediction tasks in three different inference settings, including accurately predicted structures selected with two popular criteria. The best metrics are marked by \textbf{bold}.}
\label{tab:1}
\end{table*}
Formally the D-channel positional encoding of residue pair $ij$ is denoted as
\setlength\abovedisplayskip{0.3em}
\setlength\belowdisplayskip{0.3em}
\begin{equation}
\begin{small}
\begin{aligned}
\boldsymbol{p}_{ij}=\{\mathcal{G}(\mathcal{A}(d_{ij},t_{ij};\boldsymbol{a},\boldsymbol{b}),\mu^k,\sigma^k)|k\in[1,D]\},\\
\quad\mathcal{A}(d,r;\boldsymbol{a},\boldsymbol{b})=a_rd+b_r,\cdot
\label{equ:2}
\end{aligned}
\end{small}
\end{equation}
where $\begin{aligned}\mathcal{G}(d,\mu,\sigma)=\frac1{\sigma\sqrt{2\pi}}e^{-\frac{(d-\mu)^2}{2\sigma^2}}\end{aligned}$is a Gaussian density function with parameters $\mu$ and $\sigma$, $dij$ is the Euclidean distance of residue pair $ij$, and $t_{ij}$ is the pair-type of atom pair $ij$. Please note the pair type here is not the chemical bond, and it is determined by the atom types of pair $ij$. $A(d_{ij} , t_{ij} ; a, b)$ is the affine transformation with parameters a and b; it affines $d_{ij}$ corresponding to its pair-type $t_{ij}$. Except for $d_{ij}$ and $t_{ij}$, all remaining parameters are trainable and randomly initialized. 

Next, we interactively update the sequential embedding and pairwise embedding following ~\cite{jumper2021highly, ying2021do}.
Furthermore, we adopt an orientation-aware roto-translation invariant Graphormer~\cite{luo2022antigen,jumper2021highly} to encode $\{s_i\}$ and residue frames of structure $\mathcal{P}$ into hidden representations, because the nature of the proteins will not be changed by translation or rotation and protein structure representations, should be invariant under the rotation or translation of protein structures. 
\begin{figure}[!htbp]
    \centering
    \includegraphics[width=\linewidth]{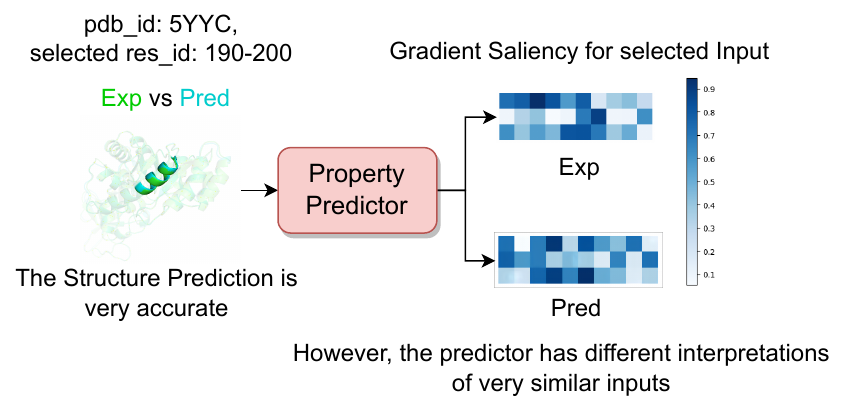}
    \vspace{-1em}
    \caption{The gradient saliency map \cite{simonyan2013deep} of target 5YYC for residue 190-200(the highlighted region). We select the first 3 dimensions for the plot}
    \vspace{-1em}
    \label{fig:2}
\end{figure}
\subsection{E. Pseudo Code and Experimental Details}
The algorithmic description of \texttt{SAO} is summarized in Algorithm \ref{alg:algorithm}.

For every downstream task, we first pretrain the encoder under \texttt{SAO} framework using the training dataset of the downstream task (\textbf{i.e., no extra data is included for every task for a fair comparison}). The hyper-parameters are summarised in the table \ref{tab:5}. The mask ratio in all pertaining is fixed as 0.15 following \cite{devlin2019bert}. Then we finetune the encoder on the training dataset of downstream tasks and test it on the test set of every downstream task following previous work~\cite{zhang2022protein}, with hyper-parameters in the table \ref{tab:6}.

All baselines and our approach are implemented using the PyTorch 1.6.0 library with Intel(R) Xeon(R) Gold6240R@2.40GHz CPU and NVIDIA V100 GPU. We select the best performance in the validation dataset and test the selected model in the test set to get the performance report.
\subsection{F. More numerical results and visualizations}
Numerical results on more tasks of the performance drop phenomena are shown in Table \ref{tab:1}. The gradient saliency map of REInet on the GO-CC dataset is shown in Figure \ref{fig:2}. The predictor predicts the GO-CC number of Target 5YYC correctly when using experimental structure while getting wrong when using highly accurate predicted structure (TMscore=0.91). The gradient saliency map shows the predictor has a very different understanding of the experimental protein structure and the very similar predicted protein structure. This gives evidence of our identified structure embedding bias.

\subsection{G. Dataset Statistic}
For the experimental protein structures, we directly use the dataset in \cite{zhang2022protein}. We retrieve the AlphaFoldDB with protein UniRef\_id for the corresponding predicted protein structure for every experimental structure. Finally, we extend the downstream tasks datasets to form experimental-predicted protein structure data pairs with labels.
\begin{table}[!htbp]
\begin{center}
\resizebox{1.0\columnwidth}{!}{
\begin{tabular}{@{}llll@{}}
\toprule
Dataset           & \# Train & \# Validation & \# Test \\ \midrule
Enzyme Commission & 15,550   & 1,729         & 1,919   \\
Gene Ontology     & 29,898   & 3,322         & 3,415   \\ \bottomrule
\end{tabular}
}
\end{center}
\vspace{-1em}
\caption{Dataset statistics for downstream tasks.}
\label{tab:3}
\end{table}

\bibliography{aaai23}